\documentclass[10pt,twocolumn,letterpaper]{article}

\usepackage{cvpr}
\usepackage{times}
\usepackage{epsfig}
\usepackage{graphicx}
\usepackage{amsmath}
\usepackage{amssymb}

\usepackage{multirow}
\usepackage{subfigure}
\usepackage{booktabs}
\usepackage{array}
\usepackage{pifont}

\usepackage[pagebackref=true,breaklinks=true,letterpaper=true,colorlinks,bookmarks=false]{hyperref}

\cvprfinalcopy % *** Uncomment this line for the final submission

\ifcvprfinal\fi

\newcommand{\tabincell}[2]{\begin{tabular}{@{}#1@{}}#2\end{tabular}}  
\makeatletter
\newcommand{\thickhline}{%
	\noalign {\ifnum 0=`}\fi \hrule height 1pt
	\futurelet \reserved@a \@xhline
}
\newcolumntype{"}{@{\hskip\tabcolsep\vrule width 1.5pt\hskip\tabcolsep}}
\makeatother
\begin{document}

%%%%%%%%% TITLE
\title{GridMask Data Augmentation}

\author{Pengguang Chen$^1$
	\quad Shu Liu$^1$ \quad Hengshuang Zhao$^1$ \quad Xingquan Wang$^2$ \quad Jiaya Jia$^1$ \\
	The Chinese University of Hong Kong$^1$ \quad City University of Hong Kong$^2$
	\\
	{\tt\small \{pgchen, sliu, hszhao, leojia\}@cse.cuhk.edu.hk \quad xingqwang3-c@my.cityu.edu.hk}
}

\maketitle
%\thispagestyle{empty}

%%%%%%%%% ABSTRACT
\begin{abstract}
	We propose a novel data augmentation method `GridMask' in this paper. It utilizes information removal to achieve state-of-the-art results in a variety of computer vision tasks. We analyze the requirement of information dropping. Then we show limitation of existing information dropping algorithms and propose our structured method, which is simple and yet very effective. It is based on the deletion of regions of the input image. Our extensive experiments show that our method outperforms the latest AutoAugment, which is way more computationally expensive due to the use of reinforcement learning to find the best policies. On the ImageNet dataset for recognition, COCO2017 object detection, and on Cityscapes dataset for semantic segmentation, our method all notably improves performance over baselines. The extensive experiments manifest the effectiveness and generality of the new method. 
\end{abstract}

%%%%%%%%% BODY TEXT
\section{Introduction}

Deep convolutional neural networks (CNNs) have achieved great success in many computer vision tasks in recent years, including image classification \cite{alexnet,vgg,inception,resnet,inceptionv2,resnext}, object detection \cite{rcnn,fast,faster,mask}, and semantic segmentation \cite{fcn, deeplab,pspnet}.
A CNN has millions of parameters, making training demand a lot of data. Otherwise, the serious over-fitting problem \cite{alexnet} could arise. Data augmentation is a very important technique to generate more useful data from existing ones for training practical and general CNNs.

\begin{table}[t]
	\centering
	\resizebox{\linewidth}{!}{
		\begin{tabular}{ l @{\hspace{0.1in}} l @{\hspace{0.1in}} l @{\hspace{0.1in}} l }
			\thickhline
			\specialrule{0em}{0pt}{3pt}		
			Task & Model & Baseline(\%) & Ours(\%) \\
			\hline
			\multirow{3}*{\tabincell{l}{Cls. \\ ImageNet}} &
			ResNet50 & 76.5 & 77.9 \textbf{(+1.4)} \\
			%		\cline{2-4}
			& ResNet101 & 78.0 & 79.1 \textbf{(+1.1)} \\
			%		\cline{2-4}
			& ResNet152 & 78.3 & 79.7 \textbf{(+1.4)} \\
			\hline
			\multirow{3}*{\tabincell{l}{Cls. \\ CIFAR10}} & ResNet18 & 95.28 & 96.54 \textbf{(+1.26)} \\
			%		\cline{2-4}
			& WideRes28-10 & 96.13 & 97.24 \textbf{(+1.11)} \\
			%		\cline{2-4}
			& Shake26-32 & 96.32 & 97.20 \textbf{(+0.88)} \\
			\hline
			\multirow{2}*{\tabincell{l}{Det. \\ COCO}} & FasterRCNN-50-FPN & 37.4 & 39.2 \textbf{(+1.8)} \\
			& FasterRCNN-X101-FPN & 41.2 & 42.6 \textbf{(+1.4)} \\
			\hline
			\tabincell{l}{Seg. \\ Cityscapes} & PSPNet50 & 77.3 & 78.1 \textbf{(+0.8)} \\
			\thickhline
		\end{tabular}
	}
	\vspace{0.1in}
	\caption{This table summarizes our results on different models and tasks. For the image classification task, we report the top-1 accuracy. For the object detection task, we report models' mAP. For the semantic segmentation task, we report models' mIoU.}
	\label{tab:t}
\end{table}

Existing data augmentation methods can be roughly divided into three categories: spatial transformation \cite{alexnet}, color distortion \cite{inception}, and information dropping \cite{cutout,randomerase,hideandseek}. Spatial transformation involves a set of basic data augmentation solutions, such as random scale, crop, flip and random rotation, which are widely used in model training.
Color distortion, which contains changing brightness, hue, etc. is also used in several models \cite{inception}.
These two methods aim at transforming the training data to better simulate real-world data, through changing some channels of information. 

Information deletion is widely employed recently for its effectiveness and/or efficiency. It includes random erasing  \cite{randomerase}, cutout \cite{cutout}, and hide-and-seek (HaS) \cite{hideandseek}. It is common knowledge that by deleting a level of information in the image, CNNs can learn originally less sensitive or important information and increase the perception field, resulting in a notable increase of robustness of the model. 

\vspace{-0.1in}
\paragraph{Motivation} Avoiding excessive deletion and reservation of continuous regions is the core requirement for information dropping methods. We found intriguingly a successful information dropping method should achieve reasonable balance between deletion and reserving of regional information on the images. The reason is twofold intuitively.

On the one hand, excessively deleting one or a few regions may lead to complete object removal and context information be removed as well. Thus remaining information is not enough to be classified and the image is more like noisy data. 
On the other hand, excessive preserving regions could make some objects untouched. They are trivial images that may lead to a reduction of the network’s robustness. Thus designing a simple method that reduces the chance of causing these two problems becomes essential.
%Overall, imbalance between deletion and reservation of continuous convex regions will lead to the failure cases, which include noisy data and trivial images. 
%Failure cases are important reasons for the failure of data augmentation, waste of computing resources and reduction of computing efficiency.

\newcommand{\flsize}{0.6in}
\begin{figure}
	\centering
	\begin{tabular}{c@{\hspace{0.03in}} c@{\hspace{0.03in}} c@{\hspace{0.03in}} c@{\hspace{0.03in}} c}
		\multicolumn{2}{c}{Cutout} & \multicolumn{2}{c}{HaS} & Ours \\
		\includegraphics[width=\flsize]{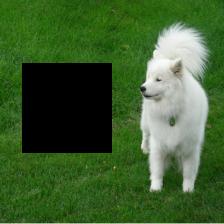} &
		\includegraphics[width=\flsize]{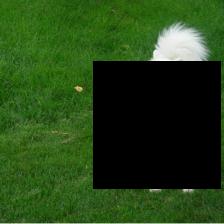} &
		\includegraphics[width=\flsize]{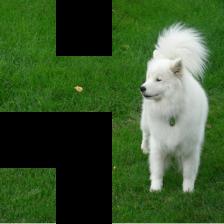} &
		\includegraphics[width=\flsize]{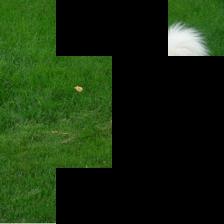} &
		\includegraphics[width=\flsize]{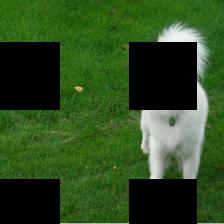} \\
		\ding{56} & \ding{56} & \ding{56} & \ding{56} & \ding{52} \\ 
	\end{tabular}
	\caption{Unsuccessful examples by previous strategies.}
	\label{fig:fail}
\end{figure}

Existing information dropping algorithms have different chances of achieving a reasonable balance between deletion and reservation of continuous regions. Both cutout \cite{cutout} and random erasing \cite{randomerase} delete {\it only one continuous} region of the image. The resulting imbalance of these two conditions is obvious because the deleted region is one area. It has a good chance to cover the whole object or none of it depending on size and location. 
%The excessive deletion on the image will cause the noisy data. If not, the excessive reservation will cause trivial images. 
The approach of HaS \cite{hideandseek} is to divide the picture evenly into small squares and delete them randomly. It is more effective and still stands a considerable chance for continuously deleting or reserving regions. 
Some unsuccessful examples of existing methods are shown in Fig. \ref{fig:fail}. Statical and more specific quantitative analysis is provided in Sec. \ref{sec:fail}.

Contrary to previous methods, we surprisingly observe the very easy strategy that can balance these two conditions statistically better is by using structured dropping regions, such as deleting uniformly distributed square regions. Our proposed information removal method, named GridMask, is thus to expand structured dropping. Its structure is really simple as shown in Fig. \ref{fig:examples}, making it easy, fast, and flexible to implement and incorporated in {\it all} existing CNN models. 

Our GridMask neither removes a continuous big region like Cutout, nor randomly selects squares like hide-and-seek. The deleted region is only a set of spatially uniformly distributed squares. In this structure, via controlling the density and size of the deleted regions, we have statistically higher chance to achieve a good balance between the two conditions, as shown in Fig.\ref{fig:balance}. As a result, we improve many state-of-the-art CNN baseline models by a good margin using our very simple GridMask at an extremely low computation budget.

To demonstrate the effectiveness of GridMask, extensive experiments are designed and conducted as shown in Table. \ref{tab:t}. In the image classification task using dataset ImageNet, GridMask can improve the accuracy of ResNet50 from 76.5\% to 77.9\%, much more effective than Cutout and HaS, which accomplish 77.1\% and 77.2\%. Our result is also better than that of AutoAugment (77.6\%), which is a combination of several existing policies through reinforcement learning. Note that our method is just one simple policy, which can also be incorporated into AutoAugment.

%Besides, other experiments on resnet101 and resnet152 are conducted, in which GridMask also increases the baseline from 78.0\% to 79.1\%, and from 78.3\% to 79.7\%, respectively. 
Further,  on the COCO2017 dataset for the object detection task, GridMask increases mAP of Faster-RCNN-50-FPN from 37.4\% to 39.2\%. On the semantic segmentation task using the challenging dataset Cityscapes, our method improves mIoU of PSPNet50 from 77.3\% to 78.1\%. They all demonstrate the surprisingly high effectiveness and generality on a large variety of tasks and training data. Our code is available at https://github.com/akuxcw/GridMask.

\newcommand{\epsize}{0.345in}
\begin{figure}[t]
	\centering
	\begin{tabular}{@{\hspace{0.0in}}c@{\hspace{0.05in}} c@{\hspace{0.1in}} c@{\hspace{0.05in}} c@{\hspace{0.1in}} c@{\hspace{0.05in}} c@{\hspace{0.1in}} c@{\hspace{0.05in}} c}
		\multicolumn{2}{c}{RErase} & \multicolumn{2}{c}{Cutout} & \multicolumn{2}{c}{HaS} & \multicolumn{2}{c}{Ours} \\
		\includegraphics[width=\epsize]{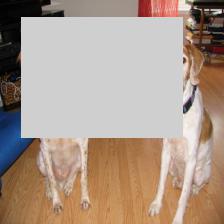} &\includegraphics[width=\epsize]{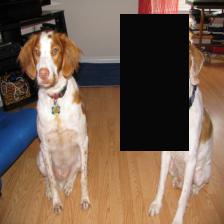} &\includegraphics[width=\epsize]{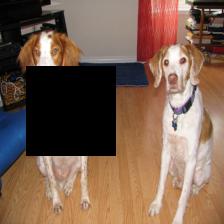} &\includegraphics[width=\epsize]{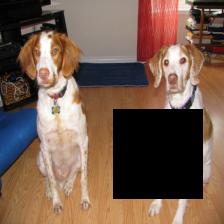} &\includegraphics[width=\epsize]{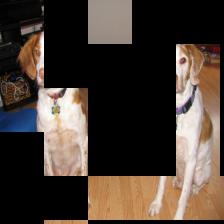} &\includegraphics[width=\epsize]{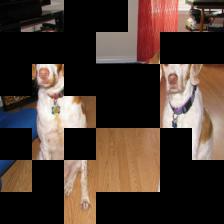} &\includegraphics[width=\epsize]{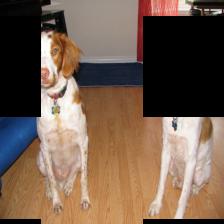} &\includegraphics[width=\epsize]{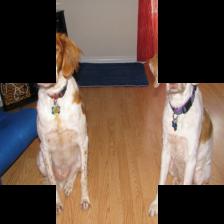} \\
		
		\includegraphics[width=\epsize]{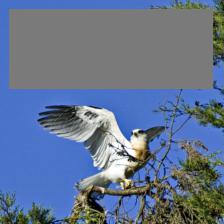} &\includegraphics[width=\epsize]{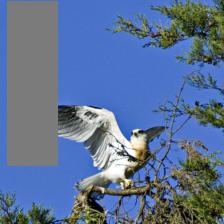} &\includegraphics[width=\epsize]{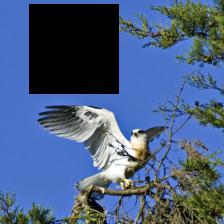} &\includegraphics[width=\epsize]{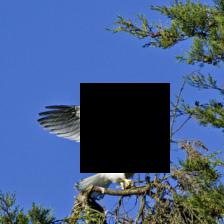} &\includegraphics[width=\epsize]{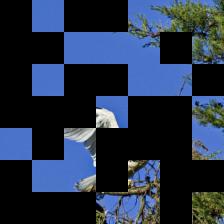} &\includegraphics[width=\epsize]{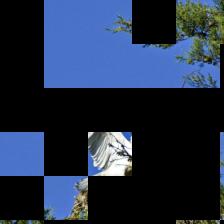} &\includegraphics[width=\epsize]{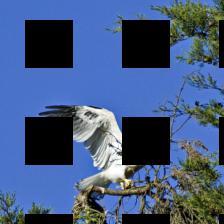} 		&\includegraphics[width=\epsize]{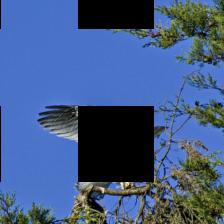} \\
		
		\includegraphics[width=\epsize]{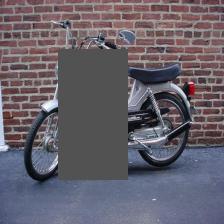}
		&\includegraphics[width=\epsize]{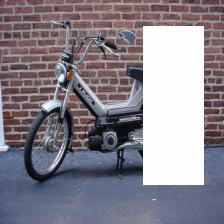}
		&\includegraphics[width=\epsize]{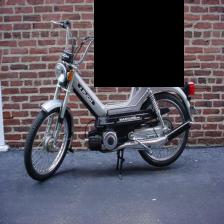}
		&\includegraphics[width=\epsize]{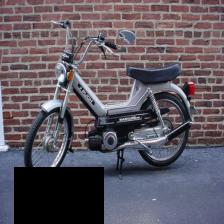}
		&\includegraphics[width=\epsize]{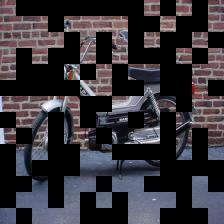}
		&\includegraphics[width=\epsize]{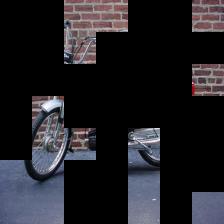}
		&\includegraphics[width=\epsize]{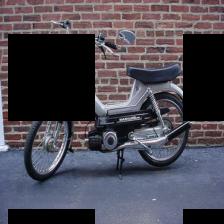}
		&\includegraphics[width=\epsize]{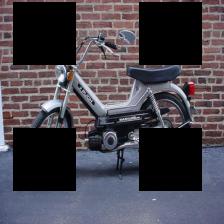} \\
		
		\includegraphics[width=\epsize]{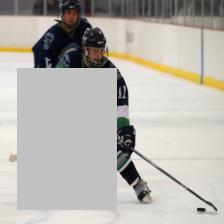}
		&\includegraphics[width=\epsize]{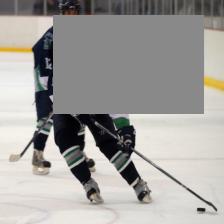}
		&\includegraphics[width=\epsize]{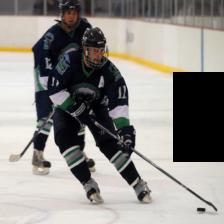}
		&\includegraphics[width=\epsize]{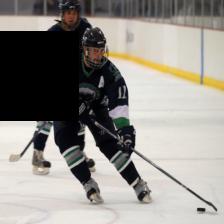}
		&\includegraphics[width=\epsize]{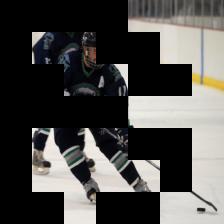}
		&\includegraphics[width=\epsize]{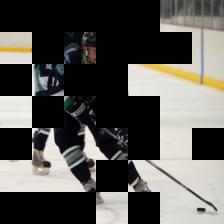}
		&\includegraphics[width=\epsize]{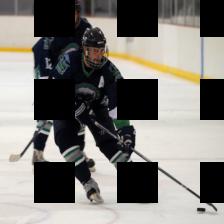}
		&\includegraphics[width=\epsize]{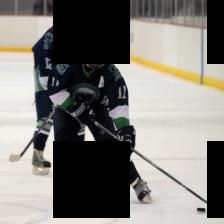} \\
		
	\end{tabular}
	\caption{More examples of different information dropping methods (best view in large size).}
	\label{fig:examples}
	
\end{figure}

\section{Releated Work}

Regularization is an important skill for training neural networks. In recent years, various regularization techniques have been proposed. Dropout \cite{dropout} is effective and is mainly used in fully connected layers.  Dropconnect \cite{dropconnect} is very similar to dropout, except that it does not drop the output value, but instead the input value of some nodes. In addition, adaptive dropout \cite{adadropout}, stochastic pooling \cite{stpool}, droppath \cite{droppath}, shakeshake regulation \cite{shakeshake} and dropblock \cite{dropblock} were also proposed. These methods add noise to a few parameters in the training process according to different rules, so as to avoid over-fitting training data and improve models' generalization ability. Besides, Mixup \cite{mixup} and CutMix \cite{cutmix} use multi-image information during the training process. By modifying the input images, labels, and loss functions, these methods can fuse information of multiple images and achieve good results.

\begin{figure}
	\centering
	\includegraphics[width=\linewidth]{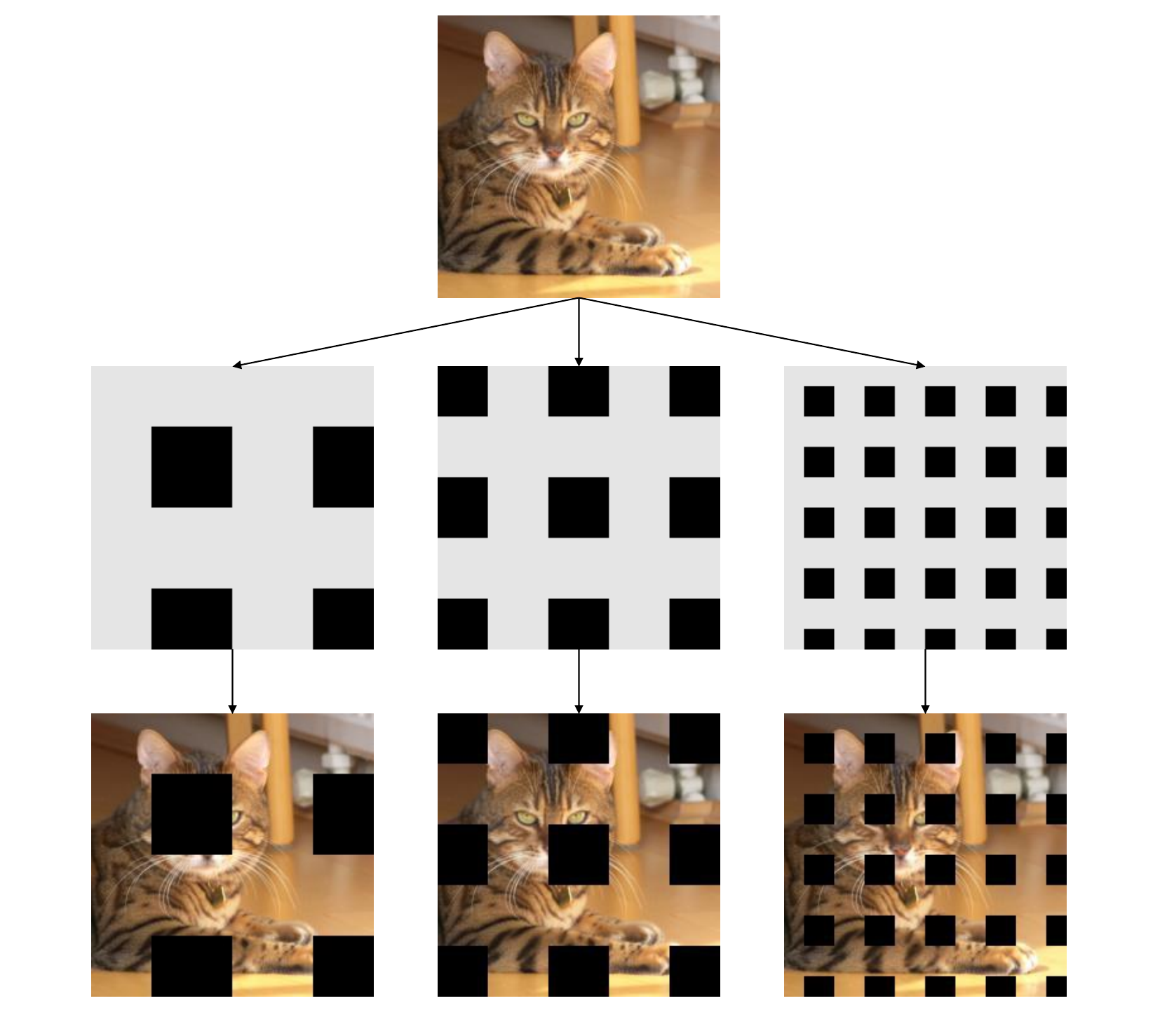}
	\caption{This image shows examples of GridMask. First, we produce a mask according to the given parameters ($r$, $d$, $\delta_x$, $\delta_y$). Then we multiply it with the input image. The result is shown in the last row. In the mask, gray value is 1, representing the reserved regions; black value is 0, for regions to be deleted.}
	\label{fig:grid}
\end{figure}

Data augmentation is also an effective regularization. 
Compared with other methods, data augmentation has many advantages. 
For example, it only needs to operate on the input data, instead of changing the network structure. 
And data augmentation is easy to apply to many tasks, while other loss- or label-based methods may need extra design. 
The basic policy of data augmentation contains random flip, random crop, etc., which are commonly used on CNNs. 
Except for the basic strategies, the inception-preprocess \cite{inception} is more advanced with random disturbance of color of the input image.
Recently, AutoAugment \cite{autoaugment} improved the inception-preprocess using reinforcement learning to search existing policies for the optimal combination. 
Besides, some recently proposed methods based on information dropping have also achieved good results of random erasing \cite{randomerase}, hide-and-seek \cite{hideandseek}, cutout \cite{cutout}, etc. 
These methods delete information on input images through certain policies. 
They usually work well on small datasets, while the effect on large datasets is limited.
 
Our method also belongs to information dropping augmentation. 
Compared with previous methods, ours can achieve consistently better results on various datasets, outperforming all previous unsupervised strategies, including the optimal combination proposed by AutoAugment. 
Our method can serve as a new baseline policy for data augmentation.

\section{GridMask}
GridMask is a simple, general, and efficient strategy. 
Given an input image, our algorithm randomly removes some pixels of it. Unlike other methods, the region that our algorithm removes is neither a continuous region \cite{cutout} nor random pixels in dropout. Instead, our algorithm removes a region with disconnected pixel sets, as shown in Fig. \ref{fig:grid}. 

We express our setting as
\begin{equation}\tilde{\textbf{x}} = \textbf{x} \times M \end{equation}%+ (1-M) \times N(\mu, \sigma) \]
where $\textbf{x} \in R^{H \times W \times C}$ represents the input image, $M \in \{0,1\}^{H \times W}$ is the binary mask that stores pixels to be removed, 
%$N(\mu, \sigma)$ generates a Gaussian noise to fill in the removed pixels in $x$, 
and $\tilde{\textbf{x}} \in R^{H \times W \times C}$ is the result produced by our algorithm. For the binary mask $M$, if $M_{i,j}=1$ we keep pixel $(i,j)$ in the input image; otherwise we remove it. Our algorithm is applied after the image normalization operation.

\begin{figure}
	\centering
	\includegraphics[width=0.4\linewidth]{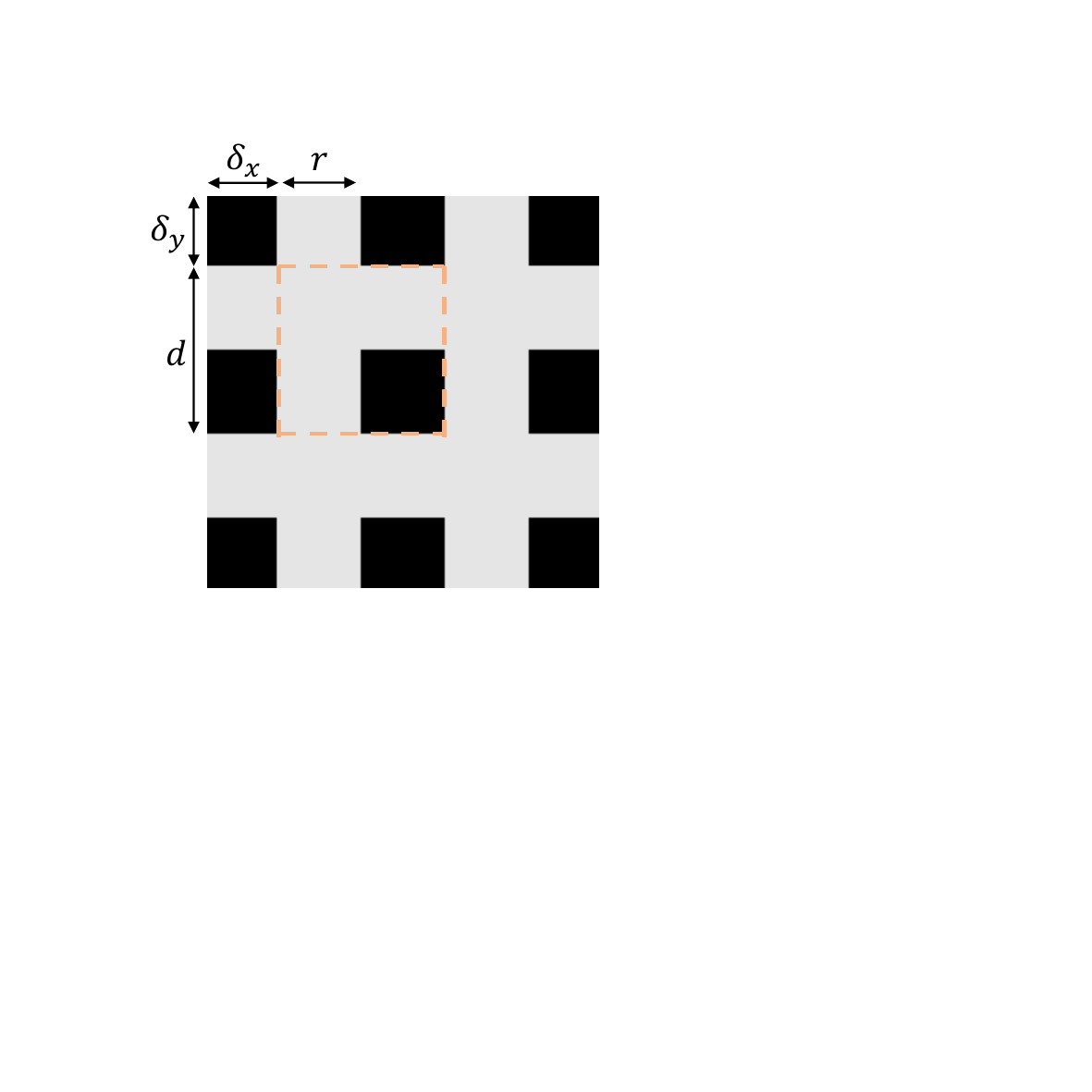}
	\caption{The dotted square shows one unit of the mask.}
	\label{fig:unit}
\end{figure}

The shape of $M$ looks like a grid, as shown in Fig. \ref{fig:grid}. We use four numbers $(r, d, \delta_x, \delta_y)$ to represent a unique $M$. Every mask is formed by tiling the units as shown in Fig. \ref{fig:unit}. $r$ is the ratio of the shorter gray edge in a unit. $d$ is the length of one unit. $\delta_x$ and $\delta_y$ are the distances between the first intact unit and boundary of the image.

Next, we talk about the choices of these four parameters. 

\paragraph{Choice of $r$}
$r$ determines the keep ratio of an input image. We define the keep ratio $k$ of a given mask $M$ as
\begin{equation}k = \frac{sum(M)}{H \times W}, \label{eq:k}\end{equation}
which means the proportion of the region between reserved and input images. The keep ratio is a very important parameter to control the algorithm. With a large keep ratio, CNN may still suffer from over-fitting. If it is too small, we could lose excessive information causing under-fitting. There is a close relation between $r$ and $k$. Ignoring incomplete units in a mask, we get
\begin{equation}k = 1 - (1-r)^2 = 2r - r^2.\end{equation}
The keep ratio is fixed following common practice. We perform extensive experimnents to verify the choice of $r$ in Section \ref{sec:abl}. 

\paragraph{Choice of $d$}
The length of one unit $d$ does not affect the keep ratio. But it decides the size of one dropped square. When $r$ is fixed, the relation between the side length $l$ of one dropped square and $d$ is
\begin{equation}l = r \times d. \end{equation}
The larger $d$ is, the larger $l$ becomes. The keep ratio is constant during training. Yet we still add randomness to enlarge the variety of images -- $d$ is suitable to achieve this goal \cite{hideandseek}. We randomly select $d$ from a range as
\begin{equation}d = random(d_{min}, d_{max}).\end{equation}
It is easy to conclude that a smaller $d$ can avoid most failure cases. But some recent works \cite{cutout, dropblock} show that dropping a very small region is useless for convolutional operation, in accordance with our experimental results given later in Section \ref{sec:abl}.

\paragraph{Choice of $\delta_x$ and $\delta_y$}
$\delta_x$ and $\delta_y$ can shift the mask given $r$ and $d$, making the mask cover all possible situations. So we randomly choose $\delta_x$ and $\delta_y$  as
\begin{equation}\delta_x(\delta_y) = random(0, d-1).\end{equation}

\paragraph{Statistics of Unsuccessful Cases}
\label{sec:fail}
Here we statistically show the probability of unsuccessful data augmentation being produced. Basically, a good balance between deletion and reservation of information is the key. We preliminary manifest that our method has lower chance to yield failure cases than Cutout and HaS.

\begin{figure}
	\centering
	\includegraphics[width=\linewidth]{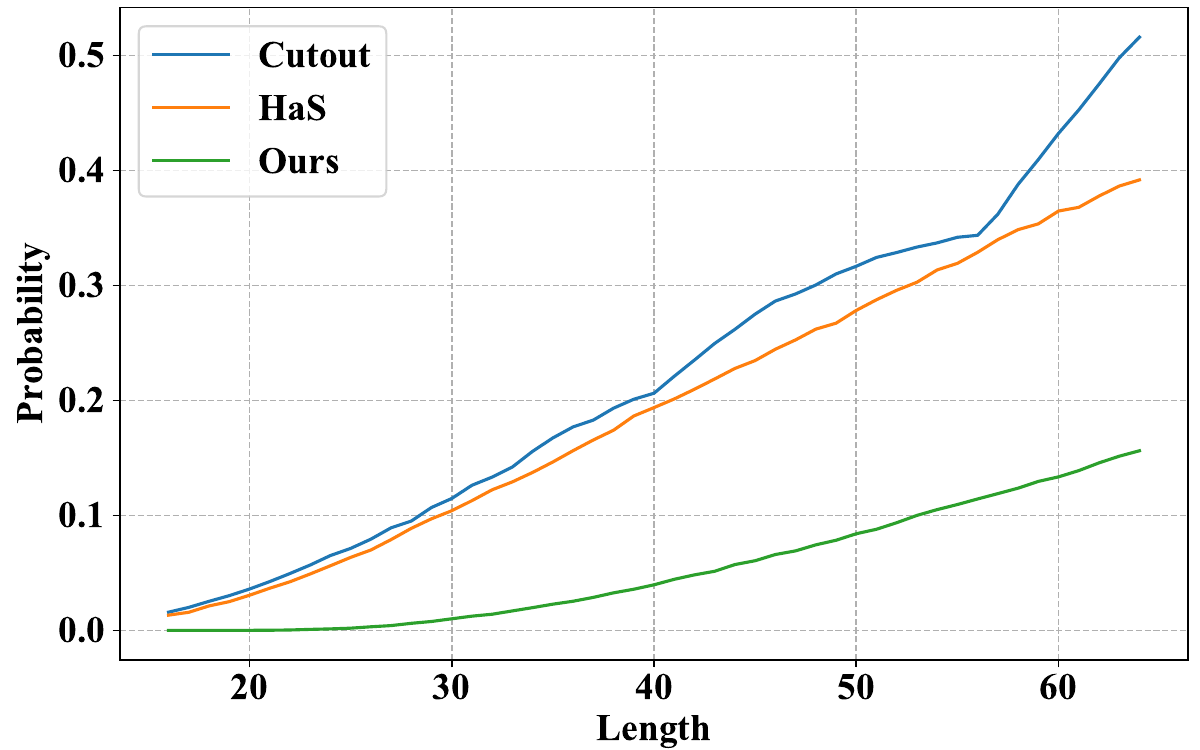}
	\caption{Statistics of failure cases with increasing of the size of dropping squares (lower probability is better). The $x$-axis shows the range of the size of one removal unit. Our method has a much lower failure probability statistically with a slower increasing trend.}
	\label{fig:balance}
\end{figure}

We simulate the condition in real datasets and calculate the probability of failure cases for different methods when varying lengths of removal squares. 
%We simulate for two kinds of datasets, one is a small-scale dataset and the other is a large-scale dataset. For the small-scale dataset, we assume all images are resized to $32 \times 32$ and the object in an image is in the size of $[12, 24]$. While for the large-scale dataset, 
All images are resized to $224 \times 224$ and the object in an image is with size within $[40, 160]$. The keep ratio is set to a typical value of 0.75 \cite{cutout} for all methods. We assume all methods randomly choose the length of removal squares within $[x, 2x]$, where $x$ is the value of the $x$-axis. Random erasing is very similar to Cutout, so we only test cutout. And we expand Cutout to multi-region Cutout for a better performance, which means randomly dropping squares until reaching the keep ratio.
If 99 percent of an object is removed or reserved, we call it a failure case.
We simulate 100,000 images and the probability of failure case for every method is summarized in Fig. \ref{fig:balance}.

Compared with other algorithms, our method always has the best performance.
With the increasing length, the superiority of our method becoming increasingly obvious. This observation allows us to choose generally larger square sizes to effectively augment data.

%When length of dropped squares is small, all methods are not easy to fail. But with increasing length, our method is much slower to cause failure examples. This observation allows us to choose generally larger square sizes to effectively augment data. 
%\jiaya{Still not clear how to define ``failure cases"!}

\vspace{-0.1in}
\paragraph{The Scheme to Use GridMask}
We use two ways to apply GridMask in practice to network training. One is to set a constant probability $p$, where we have a chance of $p$ to apply GridMask to every input image. The other way is to increase the probability of GridMask linearly with the training epochs until an upper bound $P$ is achieved. We empirically verify that the second way is better for most experiments.

\section{Experiments}
We conduct extensive experiments on several major computer vision tasks including image classification, semantic segmentation, and object detection. Our augmentation method improves the baseline on all these important tasks by a large margin. 

\newcommand{\figsize}{0.157\linewidth}
\begin{figure*}[t]
	\centering
	\begin{tabular}{@{\hspace{0.0in}} c@{\hspace{0.01in}} c@{\hspace{0.05in}} c@{\hspace{0.05in}} c@{\hspace{0.05in}} c@{\hspace{0.05in}} c@{\hspace{0.05in}} c}
		%\begin{tabular}{>{\centering\tiny}p{0.05cm}>{\tiny}c>{\tiny}c>{\tiny}c>{\tiny}c>{\tiny}c}
		& Handkerchief & Tusker & Cellphone & Pencil case & Cardigan & Fountain pen\\
		\rotatebox{90}{\ \ \ \ \quad Input} &
		\includegraphics[width=\figsize,height=\figsize]{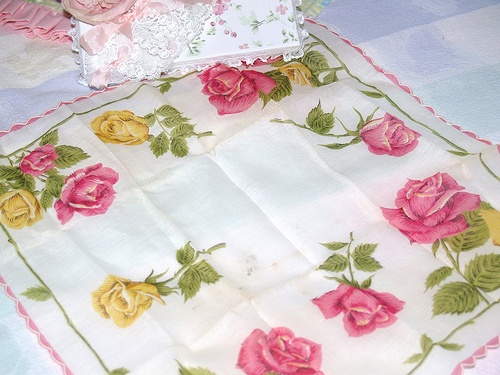} &
		\includegraphics[width=\figsize,height=\figsize]{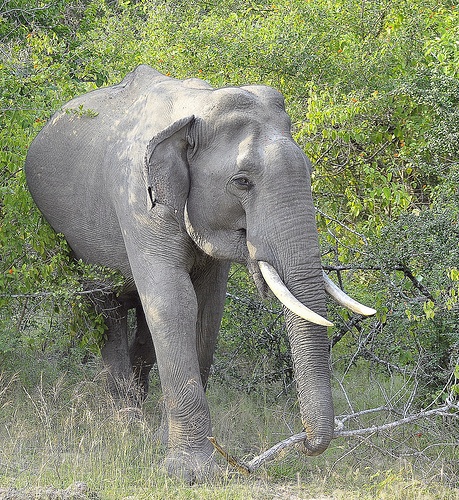} &
		\includegraphics[width=\figsize,height=\figsize]{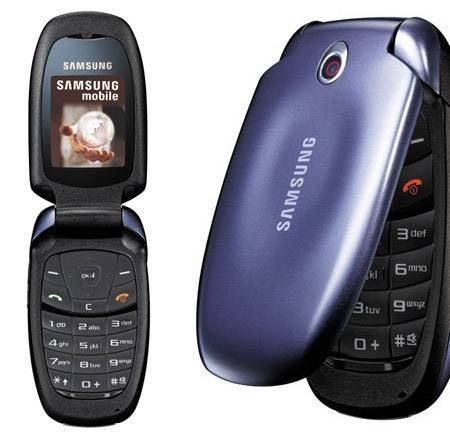} &
		\includegraphics[width=\figsize,height=\figsize]{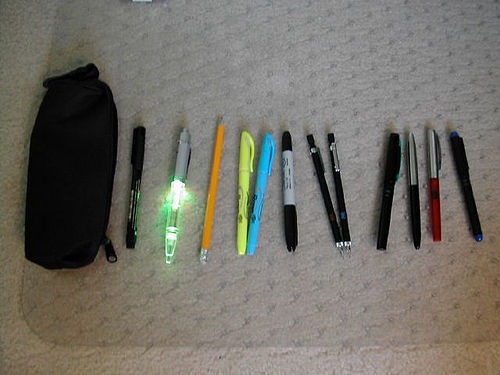} &
		\includegraphics[width=\figsize,height=\figsize]{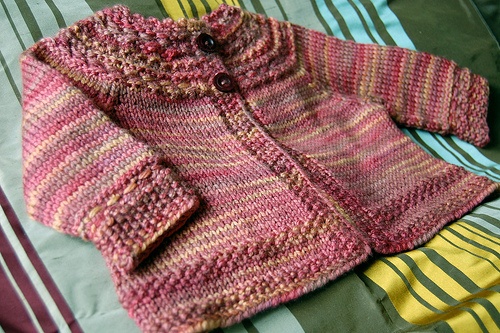} &
		\includegraphics[width=\figsize,height=\figsize]{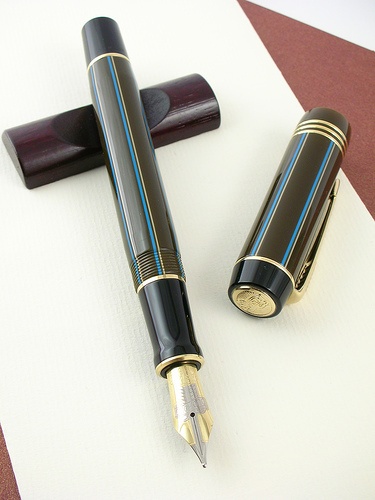} 
		\\
		
		\rotatebox{90}{\ \ \quad Baseline} &
		\includegraphics[width=\figsize,height=\figsize]{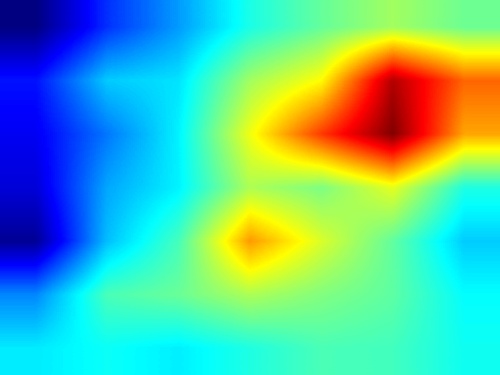} &
		\includegraphics[width=\figsize,height=\figsize]{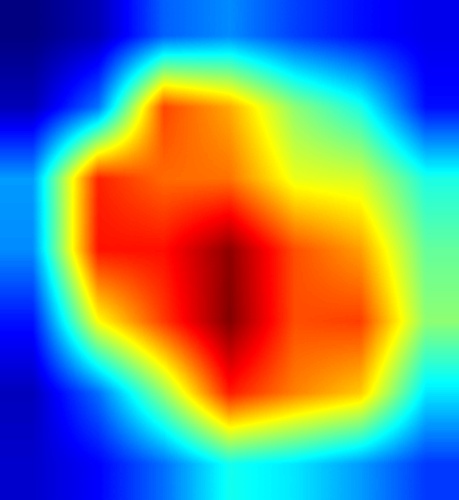} &
		\includegraphics[width=\figsize,height=\figsize]{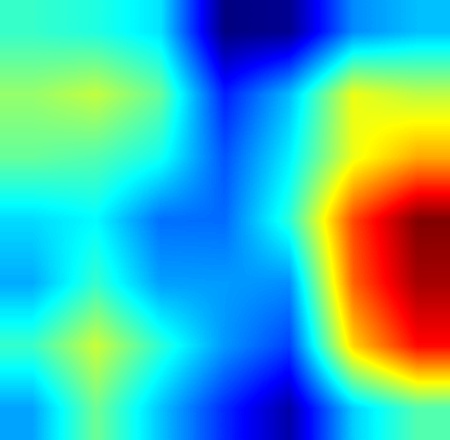} &
		\includegraphics[width=\figsize,height=\figsize]{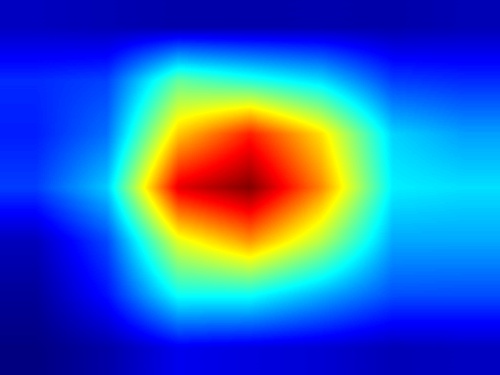} &
		\includegraphics[width=\figsize,height=\figsize]{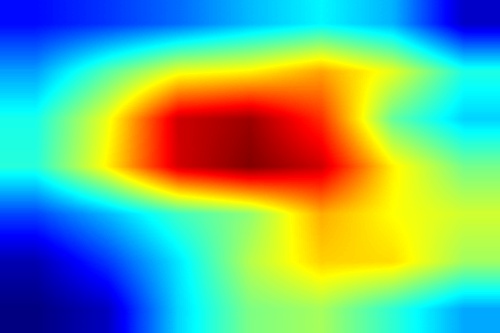} &
		\includegraphics[width=\figsize,height=\figsize]{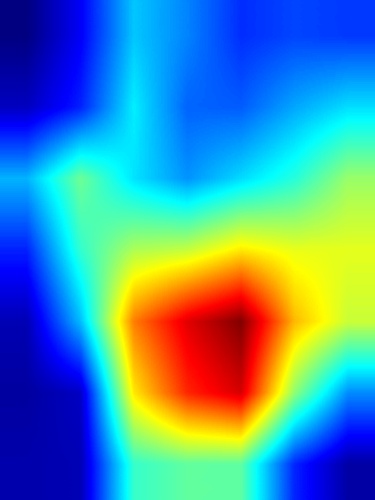}  
		\\
		
		\rotatebox{90}{\ \ AutoAugment} &
		\includegraphics[width=\figsize,height=\figsize]{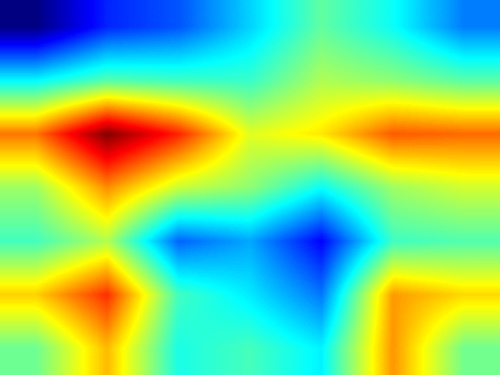} &
		\includegraphics[width=\figsize,height=\figsize]{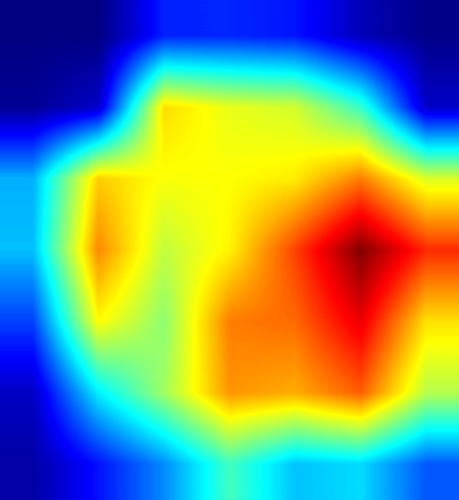} &
		\includegraphics[width=\figsize,height=\figsize]{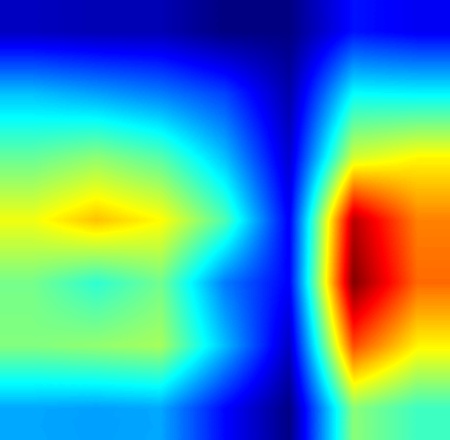} &
		\includegraphics[width=\figsize,height=\figsize]{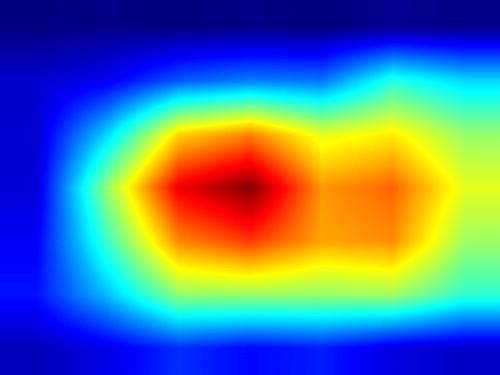} &
		\includegraphics[width=\figsize,height=\figsize]{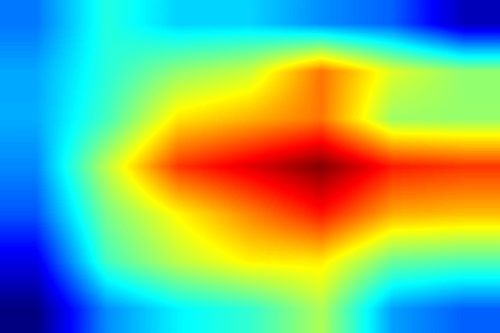} &
		\includegraphics[width=\figsize,height=\figsize]{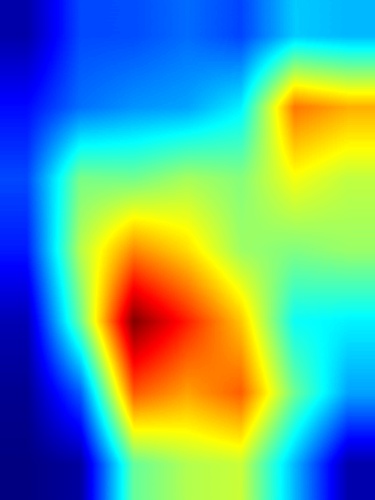} 
		\\
		
		\rotatebox{90}{\ \ \quad GridMask} &
		\includegraphics[width=\figsize,height=\figsize]{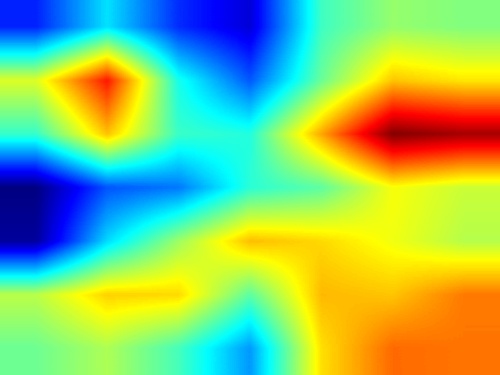} &
		\includegraphics[width=\figsize,height=\figsize]{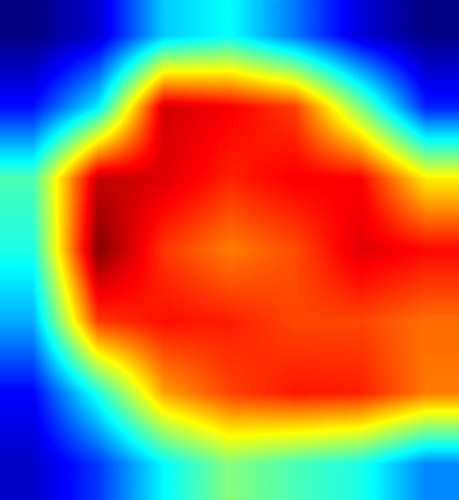} &
		\includegraphics[width=\figsize,height=\figsize]{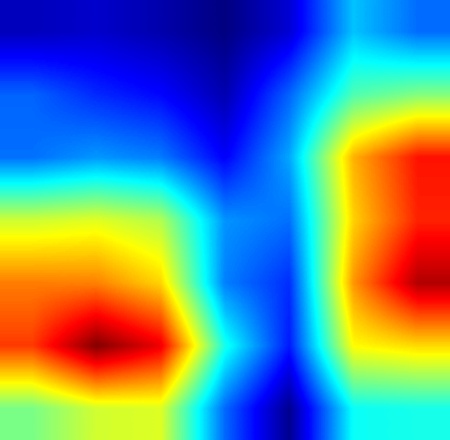} &
		\includegraphics[width=\figsize,height=\figsize]{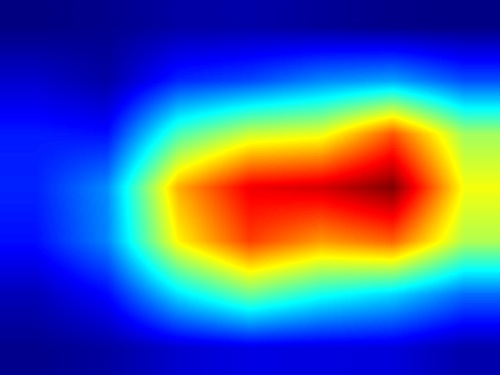} &
		\includegraphics[width=\figsize,height=\figsize]{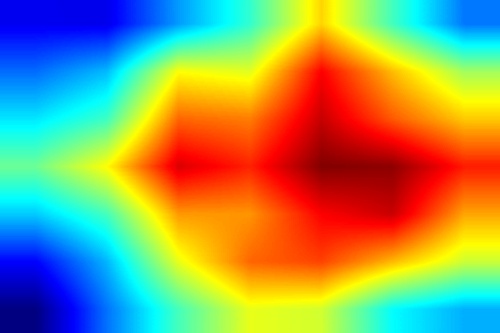} &
		\includegraphics[width=\figsize,height=\figsize]{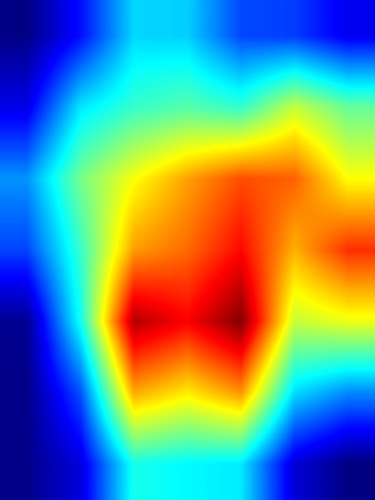} 
		\\
		
	\end{tabular}
	\vspace{0.1in}
	\caption{Class activation mapping (CAM) \cite{cam} for ResNet50 model trained on ImageNet, with baseline augment, AutoAugment or our GridMask. The models trained with AutoAugment and our strategy both incline to focus on large important regions.}
	\label{fig:cam_res}
\end{figure*}

\subsection{Image Classification}

\subsubsection{ImageNet}
\label{sec:imagenet}
ImageNet-1K is the most challenging dataset for image classification.
%, which has 1000 class, more than 1 million training images, and fifty thousand testing images. 
To demonstrate the strong capability of our proposed augmentation, we conduct experiments on it.

We experiment with a wide range of differently sized models, from ResNet50 to ResNet152. We train them with our augmentation on ImageNet from scratch for 300 epochs. The learning rate is set to 0.1, decayed by 10-fold at epochs 100, 200, 265. We train all our models on 8 GPUs, using batchsize 256 (32 per GPU). For the baseline augmentation, we follow the common practice. We first randomly crop a patch from the original image and then resize the patch to the target size (224 $\times$ 224). Finally, the patch is horizontally flipped with a probability of 0.5. 

For our method, we only use GridMask along with the baseline augmentation. We choose $r=0.6$, and we linearly increase the probability of GridMask from 0 to 0.8 with the increasing of training epochs until the 240th epoch, and then keep it until 300 epochs. The mask is also rotated before use. It is worth noting that we do not use any augmentation on color, while the strategy still consistently achieve better results, as summarized in Table \ref{tab:imagenet1}.

In terms of the accuracy, our method improves many different models, from ResNet50 to ResNet152. Our method increases ResNet50, ResNet101, and ResNet152 from 76.5\%, 78.0\%, and 78.3\% to 77.9\%, 79.1\%, and 79.7\%, respectively, with 1.4\%, 1.1\%, and 1.4\% increase. It also proves that the strategy is nicely suitable for models of various scales without careful hand tuning. 

%On ResNet50, our method outperforms popular data augmentation strategies, including Cutout and AutoAugment, and achieves the best results. \jiaya{Explain why only resnet50? How about other models?}

\begin{table}[t]
	\centering
	\begin{tabular} { p{2.1in}  p{0.7in}<{\centering} }
		\thickhline
		\specialrule{0em}{0pt}{2pt}
		Model & Accuracy(\%) \\
		\hline
		ResNet50 \cite{dropblock} & 76.5  \\
		ResNet50 + Cutout \cite{cutmix} & 77.1 \\
		ResNet50 + HaS \cite{hideandseek} & 77.2 \\
		ResNet50 + AutoAugment \cite{autoaugment} & 77.6 \\
		ResNet50 + GridMask (Our Impl.) & \textbf{77.9} \\
		\hline
		ResNet101 \cite{mixup} & 78.0 \\
		ResNet101 + GridMask (Our Impl.) & \textbf{79.1} \\
		\hline
		ResNet152 \cite{resnet} & 78.3 \\
		ResNet152 + GridMask (Our Impl.) & \textbf{79.7} \\
		\thickhline
	\end{tabular}
	\vspace{0.1in}
	\caption{This table summarizes the result of ImageNet. We can see our model improves the baseline of various models.}
	\label{tab:imagenet1}
\end{table}

\paragraph{Comparison with Cutout} 
Cutout \cite{cutout} also does not distort color and only drops information. Its performance on ImageNet improves ResNet50 by 0.6\% (from 76.5\% to 77.1\%). Our method drops information in a more effective structure, improving ResNet50 by 1.4\% on ImageNet. 

\paragraph{Comparison with HaS} 
HaS \cite{hideandseek} is the previous SOTA information dropping method, which is better than Cutout. It uses smaller removal squares (between sizes 16 and 56). When the squares get larger, the result becomes worse in the experiments reported in the original paper. 
Our setting, contrarily, produces better results even when removal squares are large.
It is because we handle the aforementioned failure cases better.

\paragraph{Comparison with AutoAugment} AutoAugment \cite{autoaugment} is a SOTA data augmentation method. It uses reinforcement learning to search using tens of thousands of GPU hours to find a combination of existing augmentation policies. It thus performs reasonably better than previous strategies and improve the accuracy of ResNet50 by 1.1\%. Our method, by simply dropping part of the information of the input image in a regular way, even exceeds the effect of AutoAugment. Our method is extremely easy, only uses one data augmentation policy, and still performs more satisfyingly than this type of exhaustive combination of various data augmentation policies. The effectiveness and generality are well demonstrated.

\paragraph{Benefit to CNN} To analyze what the model trained with our GridMask learns, we compute class activation mapping (CAM) \cite{cam} for ResNet50 model trained with our policy on ImageNet. We also show the CAM for models trained with baseline augmentation and AutoAugment for comparison. We intriguingly observe common properties between our method and AutoAugment. Compared to the baseline method, both AutoAugment and ours tend to focus on larger spatially distributed regions. It indicates successful augmentation makes the system put attention to large and salient representations. It can quickly improve the generalization ability of models. This figure also demonstrates that the method with our strategy attends to most structurally comprehensive regions. The third image is an example, where the two cellphones are both important. The baseline method just focuses on the right phone, and AutoAugment pays attention to the left one. Contrarily, our method notices both cellphones and helps recognition with this set of more accurate information.

\begin{table}[t]
	\centering
	\begin{tabular} { p{2.3in} @{\hspace{0.0in}} p{0.8in}<{\centering} }
		\thickhline
		\specialrule{0em}{0pt}{2pt}
		Model & Accuracy (\%) \\
		\hline
		ResNet18 \cite{resnet} & 95.28 \\
		~ + Randomerasing \cite{randomerase} & 95.32 \\
		~ + Cutout \cite{cutout} & 96.25 \\
		~ + HaS \cite{hideandseek} & 96.10 \\
		~ + AutoAugment \cite{autoaugment} & 96.07 \\
		~ + GridMask (Ours) & 96.54 \\
		\hline
		~ + AutoAugment \& Cutout \cite{autoaugment} & 96.51 \\
		~ + AutoAugment \& GridMask (Ours) & \textbf{96.64} \\
		\hline
		\hline
		WideResNet-28-10 \cite{wideresnet} & 96.13 \\
		~ + Radnomerasing \cite{randomerase}* & 96.92 \\
		~ + Cutout \cite{cutout} & 97.04 \\
		~ + HaS \cite{hideandseek} & 96.94 \\
		~ + AutoAugment \cite{autoaugment} & 97.01 \\
		~ + GridMask (Ours) & 97.24 \\
		\hline
		~ + AutoAugment \& Cutout \cite{autoaugment} & 97.39 \\
		~ + AutoAugment \& GridMask (Ours) & \textbf{97.48} \\
		\hline	
		\hline
		ShakeShake-26-32 \cite{shakeshake} & 96.42 \\
		~ + Randomerasing \cite{randomerase} & 96.46 \\
		~ + Cutout \cite{cutout} & 96.96 \\
		~ + Has \cite{hideandseek} & 96.89 \\
		~ + Autoaugment \cite{autoaugment} & 96.96 \\
		~ + GridMask (Ours) & 97.20 \\
		\hline
		~ + AutoAugment \& Cutout \cite{autoaugment} & 97.36 \\
		~ + AutoAugment \& GridMask (Ours) & \textbf{97.42} \\
		\thickhline
	\end{tabular}
	\vspace{0.1in}
	\caption{Results on CIFAR10 are summarized in this table. We achieve the best accuracy on different models. * means results reported in the original paper.}
	\label{tab:cifar10}
\end{table}

\subsubsection{CIFAR10}
The CIFAR10 dataset has 50,000 training images and 10,000 testing images. CIFAR10 has 10 classes, each has 5,000 training images and 1,000 testing images. 

We summarize the result on CIFAR10 in Table \ref{tab:cifar10}. We use open-source PyTorch implementations for ResNet-18, WideResNet-28-10, and Shake-Shake. We use the same hyperparameters as reported in the paper \cite{cutout,shakeshake}, except using larger training epochs for ResNet-18 and WideResNet-28-10. We use the same hyperparameters to train all methods. For the baseline augmentation, we first pad the input to 40$\times$40 and randomly crop a patch of size 32$\times$32. Depending on models, the patch is chosen to be horizontally flipped or not. Other augmentation methods are added after the baseline augmentation. We use $r=0.4$, and the same scheduling method as described in Section \ref{sec:imagenet}. One thing to note is that, in \cite{autoaugment}, authors train their policies together with Cutout. For the sake of fairness, we add Cutout and AutoAugment separately in our experiments. Some results are our reimplementation with the same training strategy as ours -- we achieve similar results reported in the original papers. We train every network for three times and report the mean accuracy.

The table indicates that our GridMask improves many baseline models by a large margin. With GridMask, we improve the accuracy of ResNet18 from 95.28\% to 96.54\% (+1.26\%), which surpasses previous information dropping methods significantly. Also, our result is better than AutoAugment. For other models, our method can improve the accuracy of WideResNet28-10 and ShakeShake-26-32 from  96.13\% and 96.43\% to 97.24\% (+1.11\%), and 97.20\% (+0.88\%), which is still better than other augmentation methods. Combined with AutoAugment, we achieve SOTA result on these models.

\subsubsection{Ablation Study}
\label{sec:abl}
In this section, we train models with GridMask under different parameters and show variations of GridMask. 

\begin{figure}[t]
\subfigure[ImageNet]{
	\begin{minipage}[t]{\linewidth}
		\centering
	\includegraphics[width=0.6\linewidth]{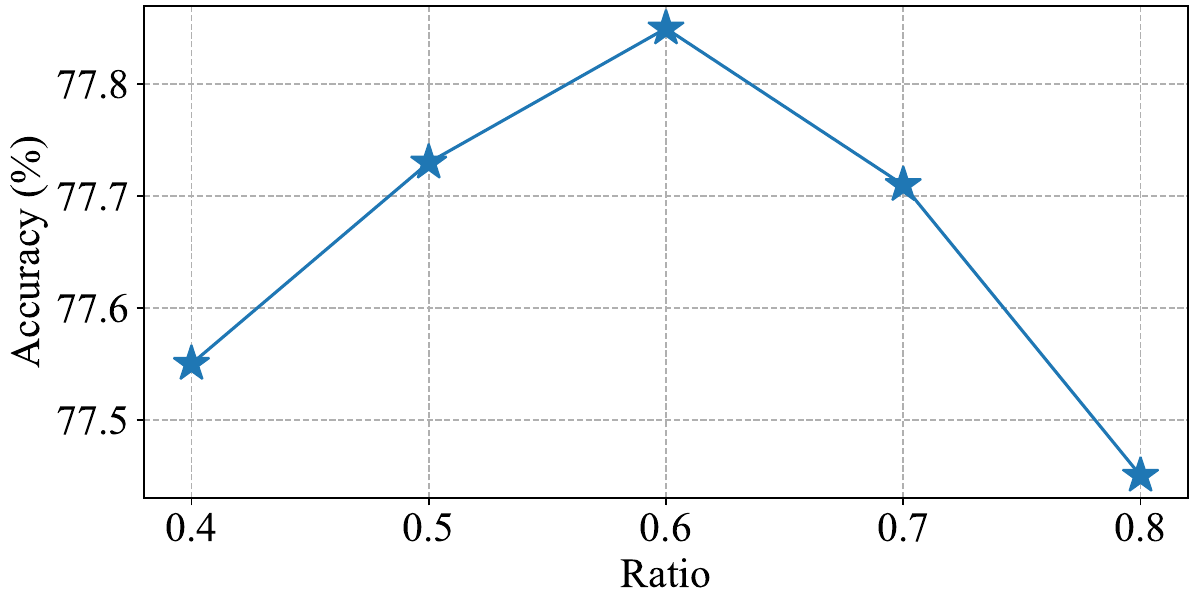}
\label{fig:ratio}	
\end{minipage}
}

\subfigure[CIFAR10]{
	\begin{minipage}[t]{\linewidth}
		\centering
		\includegraphics[width=0.6\linewidth]{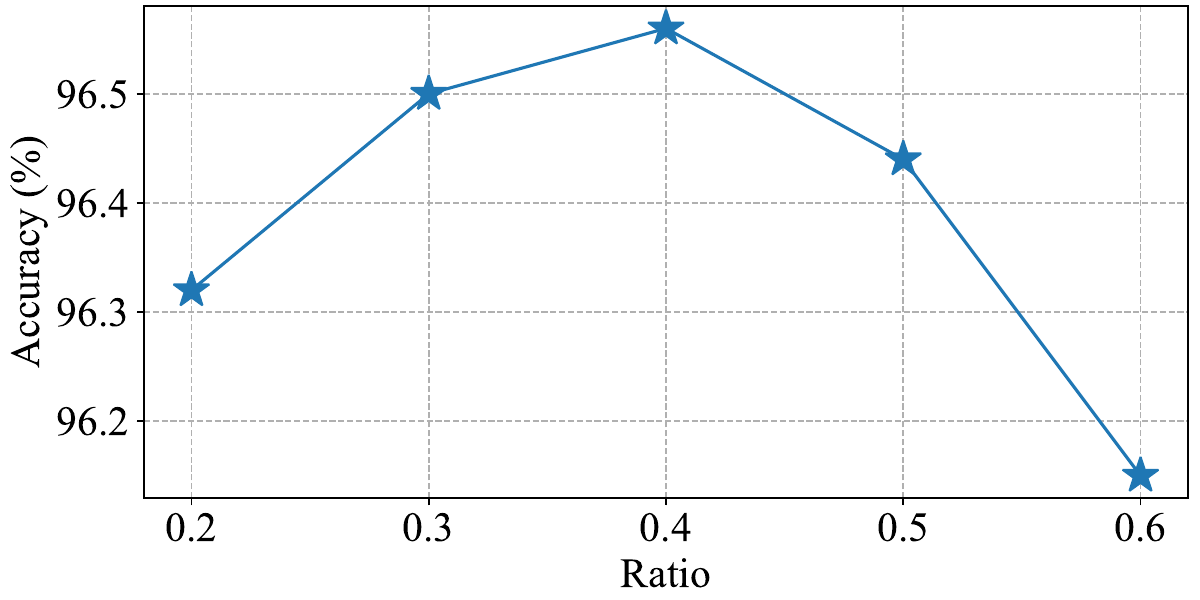}
\label{fig:ratio2}		
	\end{minipage}
}
	\centering
	\caption{The accuracies of different $r$ on ImageNet and CIFAR10.}

\end{figure}

\paragraph{Hyperparameter $r$} We experiment with setting $r$ as 0.4, 0.5, 0.6, 0.7, and 0.8 on ImageNet with ResNet50. The result is summarized in Fig. \ref{fig:ratio}. According to the result, we choose the most effective $r = 0.6$ as the choice of $r$ for different models on the Imagenet dataset. We also experiment with $r$ being 0.2, 0.3, 0.4, 0.5, and 0.6 on CIAFR10 with ResNet18. The result is shown in Fig. \ref{fig:ratio2}, and we choose $r = 0.4$ as the choice of $r$ on CIFAR10. Through experiments, it is important to note that the $r$ selected on more complex datasets (such as ImageNet) becomes larger. Put differently, we should keep more information on complex datasets to avoid under-fitting, and delete more on simple datasets to reduce over-fitting. This finding is in obedience to our common sense.

\begin{table}[t]
	\centering
	\begin{tabular}{ p{1.2in} p{1.3in}<{\centering} }
		\thickhline
		\specialrule{0em}{0pt}{2pt}
		Range of $d$ & Accuracy (\%) \\
		\hline
		$[40, 60]$ & 77.26 \\
		$[96, 120]$ & 77.58 \\
		$[150, 170]$ & 77.61 \\
		$[200, 224]$ & 77.57 \\
		\textbf{[96, 224]} & \textbf{77.89} \\
		\thickhline
	\end{tabular}
	\vspace{0.05in}
	\caption{Results on different ranges of $d$. }
	\label{tab:d}
\end{table}

\paragraph{Hyperparameter $d$} We experiment with setting different ranges of $d$ on ImageNet, and the results are summarized in Table \ref{tab:d}. When the range of $d$ is concentrated in some small intervals, the accuracy is low. Also when $d$ is too small, the result is even worse. We set $d$ in the optimal range of $[96,224]$. These experiments verify our previous analysis that different $d$ can bring varying effect to networks, and the diversity of $d$ can increase robustness of the network.

\paragraph{Variations of GridMask} The first variation is reversed GridMask, which means we keep what we drop in GridMask, and drop what we keep in GridMask. According to our analysis, the reversed GridMask should yield similar performance on different challenging datasets because a good balance of the two conditions in GridMask should be similarly good for reserved GridMask. We try different $r$ for reversed GridMask. The result is listed in Table \ref{tab:rev}. The reversed GridMask runs better than other augmentation methods. 

\begin{table}[ht]
	\centering
	\begin{tabular}{c  l  c}
		\thickhline
		\specialrule{0em}{0pt}{2pt}		
		Dataset & Model & Accuracy(\%) \\
		\hline
		\multirow{4}*{ImageNet} & ResNet50 & 76.51 \\
		& ~ + RevGridMask ($r=0.1$) & 77.42 \\
		& ~ + RevGridMask ($r=0.2$) & 77.74 \\
		& ~ + RevGridMask ($r=0.3$) & 77.55 \\
		\hline
		\multirow{4}*{CIFAR10} & ResNet18 & 95.28 \\
		& ~ + RevGridMask ($r=0.2$) & 96.18 \\
		& ~ + RevGridMask ($r=0.3$) & 96.46 \\
		& ~ + RevGridMask ($r=0.4$) & 96.33 \\
		\thickhline
	\end{tabular}
	\vspace{0.1in}
	\caption{Results of reversed GridMask. We still get good results, proving the robustness of our algorithm and superiority of structured information dropping methods.}
	\label{tab:rev}	
\end{table}

Another variation of GridMask is random GridMask. 
In the GridMask, we can regard the mask as composed of many units, and we drop a block in every unit. 
This forms our structured information dropping. 
A natural variation is to break the structure and randomly drop a block in every unit with a certain probability of $p_u$. 
The result is summarized in Table \ref{tab:random}. 
Using random dropping decreases the final accuracy. 
The structured information dropping is more effective.

\begin{table}[t]
	\centering
	\begin{tabular} {p{0.8in}  p{0.4in}<{\centering}  p{0.4in}<{\centering}  c }
		\thickhline
		\specialrule{0em}{0pt}{1pt}
		Model & $r$ & $p_u$ & Accuracy (\%) \\
		\hline
		\multirow{9}*{ResNet18} & \multirow{3}*{0.3} & 0.5 & 96.43 \\
		& & 0.7 & 96.35 \\
		& & 0.9 & 96.40 \\
		\cline{2-4}
		& \multirow{3}*{0.4} & 0.5 & 96.13 \\
		& & 0.7 & 96.37 \\
		& & 0.9 & 96.42 \\
		\cline{2-4}
		& \multirow{3}*{0.5} & 0.5 & 96.05 \\
		& & 0.7 & 96.39 \\
		& & 0.9 & 96.40 \\
		\thickhline
	\end{tabular}
	\vspace{0.1in}
	\caption{Results of random GridMask. The statistics show randomly drop information does not help improve the results.}
	\label{tab:random}
\end{table}

\subsection{Object Detection on COCO Dataset}

In this section, we use our GirdMask policy to train objection detectors on the COCO dataset, to show our method is a generic augmentation policy. We use Faster-RCNN as our baseline model with open-source PyTorch implementation \cite{massa2018mrcnn}. All models are initialized using an ImageNet pre-trained weight and are then finely tuned for some epochs on the COCO2017 dataset.

The baseline augmentation including randomly deforming the brightness, contrast, saturation, and hue of the input image. And then the image is randomly scaled into a certain range. After that, a horizontal flip operation is randomly applied to the scaled image. Finally, the image is normalized to around zero. Our GridMask is used after the baseline augmentation. 

We use the same hyperparameters as described in \cite{massa2018mrcnn}, except for the training epochs. We first double the original training epochs for both baseline and our GridMask. Then, we increase the training time for both methods further -- but the baseline models face a serious over-fitting problem and tend to decrease after $2\times$ training epochs. But models trained with our GridMask yield better results. This demonstrates that our method can handle the over-fitting problem generally and essentially.

The result of our method with different hyperparameters is shown in Table \ref{tab:det}. We choose $r = 0.5$ following previous experience. The experiments with different probability $p$ on Faster-RCNN-50-FPN are conducted. With a large range of $p$ from 0.3 to 0.9, we all achieve excellent results, which only fluctuate between 38.0\% and 38.3\%. This further demonstrates the stability of our method. When the probability $p$ is 0.7, we obtain the best result, which increases mAP by 0.9\%. When we further increase the training epochs, we get a higher result of 39.2\%, which promotes the baseline by 1.8\%. For Faster-RCNN-X101-32x8d-FPN, we increase the mAP from 41.2\% to 42.6\%, by 1.4\%.

\begin{table}[t]
	\centering
	\resizebox{\linewidth}{!}{	
		\begin{tabular}{ l @{\hspace{0.1in}} c @{\hspace{0.1in}} c @{\hspace{0.1in}} c}
			\thickhline
			\specialrule{0em}{0pt}{2pt}
			Model & mAP (\%) & AP50 (\%) & AP75 (\%) \\
			\hline
			Faster-RCNN-50-FPN ($2\times$) & 37.4 & 58.7 & 40.5 \\
			~ + GridMask ($p = 0.3$)  & 38.2 & 60.0 & 41.4 \\
			~ + GridMask ($p = 0.5$)  & 38.1 & 60.1 & 41.2 \\
			~ + GridMask ($p = 0.7$)  & \textbf{38.3} & \textbf{60.4} & \textbf{41.7} \\
			~ + GridMask ($p = 0.9$)  & 38.0 & 60.1 & 41.2 \\
			\hline
			Faster-RCNN-50-FPN ($4\times$)  & 35.7 & 56.0 & 38.3 \\
			~ + GridMask ($p = 0.7$)  & \textbf{39.2} & \textbf{60.8} & \textbf{42.2} \\
			\hline
			\hline
			Faster-RCNN-X101-FPN ($1\times$) & 41.2 & 63.3 & 44.8 \\
			Faster-RCNN-X101-FPN ($2\times$) & 40.4 & 62.2 & 43.8 \\
			~ + GridMask ($p = 0.7$) & \textbf{42.6} & \textbf{65.0} & \textbf{46.5} \\
			\thickhline
		\end{tabular}
	}
	\vspace{0.1in}
	\caption{Result of object detection on the COCO2017 dataset. Our method improves Faster-RCNN-50-FPN and Faster-RCNN-X101-32x8d-FPN significantly, by 1.8\% and 1.4\%.}
	\label{tab:det}
\end{table}

\subsection{Semantic Segmentation on Cityscapes}

Semantic segmentation is a challenging task in computer vision, which densely predicts the semantic category for every pixel in an image. To demonstrate the universality of our GridMask, we also conduct experiments on challenging Cityscapes dataset. 

We use PSPNet \cite{pspnet} as our baseline model, which achieved SOTA results for semantic segmentation. We use the same hyperparameters as suggested in \cite{semseg2019}, except for the training epochs. We train for longer epochs following common practice. We do not increase the training epochs for the baseline model, because training longer will cause serious over-fitting problems and decrease the accuracy of the baseline model. All models are initialized by the same ImageNet pre-trained weights and then fine-tuned on the Cityscapes dataset.

\begin{table}[ht]
	\centering
	\begin{tabular}{p{1.0in} p{0.7in} p{0.9in}<{\centering}}
		\thickhline
		\specialrule{0em}{0pt}{2pt}
		model & Method & mIoU (\%) \\
		\hline
		\multirow{2}*{PSPNet50} & Baseline & 77.3 \\
		& GridMask & \textbf{78.1} \\
		\hline
		\multirow{2}*{PSPNet101} & Baseline & 78.6 \\
		& GridMask & \textbf{79.0} \\
		\thickhline
	\end{tabular}
	\vspace{0.1in}
	\caption{Result of semantic segmentation on the Cityscapes dataset. We train our models on fine set and report mIoU on the validation set.}
	\label{tab:cityscapes}
\end{table}

The baseline model already uses strong augmentation policies, including randomly scaling the image from 0.5 to 1.0, randomly rotating the image in $\pm 10$ degrees, with random Gaussian blur the image, with random horizontal flip of the image, and randomly cropping a patch form the image. The strong baseline augmentation greatly raises the difficulty of adding another augmentation policy. Surprisingly, we still achieve a better result after adding our GridMask along with the baseline augmentation. We summarize the result in Table \ref{tab:cityscapes}.

\subsection{Expand Grid as Regularization}

Data augmentation is only one of the regularization methods. Shape grid is not only useful in data augmentation but also work in other aspects. Inspired by \cite{cutmix}, we combine our Grid shape with Mixup. And we train ResNet50 with our method on ImageNet, we also obtain SOTA results compared with other regularization methods, as shown in Table \ref{tab:imagenet2}.

\begin{table}[t]
	\centering
	\begin{tabular} {p{1.7in} p{1.1in}<{\centering}}
		\thickhline
		Method & Accuracy (\%) \\
		\hline
		ResNet50 \cite{dropblock} & 76.5 \\
		~ + Dropout \cite{dropblock} & 76.8 \\
		~ + Label sommthing \cite{dropblock} & 77.2 \\
		~ + Mixup \cite{cutmix} & 77.4 \\
		~ + DropBlock \cite{dropblock} & 78.1 \\
		~ + CutMix \cite{cutmix} & 78.6 \\
		~ + Ours & \textbf{78.7} \\
		\thickhline
	\end{tabular}
	\vspace{0.1in}
	\caption{Combining our GridMask with Mixup, we achieve the best result in all regularization methods.}
	\label{tab:imagenet2}
\end{table}

\section{Discussion and Conclusion}
We have proposed a simple, general, and effective policy for data augmentation, which is based on information dropping. It deletes uniformly distributed areas and finally forms a grid shape. Using this shape to delete information is more effective than setting  complete random location. It has achieved remarkable improvement in different tasks and models. On the ImageNet dataset, it increases the baseline by 1.4\%. In the task of COCO2017 object detection, we improve the baseline by 1.8\%, and in the task of Cityscapes semantic segmentation, we boost the baseline by 0.9\%. This effect is consistently stronger than other information deletion based data augmentation methods. Further, our method can serve as a new baseline policy in future data augmentation search algorithms. 

Our method is one successful way of using structured information dropping, and we believe there are more also with excellent structures. We hope the study on information dropping methods inspires more future work to understand the importance of designing effective structures, which may even help reinforcement learning to get improved.

{\small
\bibliographystyle{ieee_fullname}
\bibliography{egbib}
}

\end{document}